\documentclass[conference]{IEEEtran}

\IEEEoverridecommandlockouts                              

\usepackage{amsmath,amssymb,graphics,epsfig,xcolor,tikz,siunitx,booktabs,pgfplots,grffile}
\usepackage[caption=false]{subfig}
\usepackage{algorithm}
\usepackage[noend]{algpseudocode}
\pgfplotsset{compat=newest}
\usetikzlibrary{plotmarks}
\usetikzlibrary{arrows.meta}
\usepackage[keeplastbox]{flushend}
\usepackage[style=ieee,backend=biber]{biblatex}
\AtEveryBibitem{%
  \clearfield{issn} 
  \clearfield{isbn} 
  \clearfield{doi} 
  \clearfield{month}
  \ifentrytype{online}{}{
    \clearfield{url}
  }
}

\begin{document}

\title{Autonomous Vehicle Speed Control for Safe Navigation of Occluded Pedestrian Crosswalk}

\author{Sarah M. Thornton$^{*}$
\thanks{$^{*}$S.M. Thornton is with the Dynamic Design Lab in the Department of Mechanical Engineering, Stanford University, Stanford, CA, 94305 USA. \mbox{E-mail}: smthorn@stanford.edu}%
}

\maketitle
\thispagestyle{plain}
\pagestyle{plain}

\begin{abstract}
Both humans and the sensors on an autonomous vehicle have limited sensing capabilities. When these limitations coincide with scenarios involving vulnerable road users, it becomes important to account for these limitations in the motion planner. For the scenario of an occluded pedestrian crosswalk, the speed of the approaching vehicle should be a function of the amount of uncertainty on the roadway. In this work, the longitudinal controller is formulated as a partially observable Markov decision process and dynamic programming is used to compute the control policy. The control policy scales the speed profile to be used by a model predictive steering controller.
\end{abstract}

\section{Introduction}\label{sec:intro}
Autonomous vehicles rely on sensors to provide information about the world to decision-making algorithms. Just like humans have limited sensing capabilities, the sensors on an autonomous vehicle are also susceptible to limitations. GPS requires open skies. Cameras require certain weather and lighting conditions. Radar, while less affected by weather, has limited resolution, range and field of view. Lidar uses lasers to improve resolution, but also has limited range and field of view. Even when these sensors operate ideally, they cannot ``see" everything. For example, when a vehicle is driving behind a larger vehicle, such as a semi-truck, part of the environment is occluded both to the human occupants and the sensor suite. Sensor occlusion increases uncertainty in the environment. If not properly accounted for, it can unintentionally allow decision-making algorithms to operate at risky conditions. When combined with a scenario where vulnerable road users are hidden by the occlusions, it can lead to unsafe (and potentially illegal) navigation of the roadways. 

An autonomous vehicle equipped with lidar sensors would have a \SI{360}{\degree} 3D point cloud (x,y,z tuples) of surface distances relative to the vehicle. Many vehicle control and decision-making algorithms only account for planar motion of the vehicle \cite{Brown2016}, thus, rendering most of the 3D information superfluous. To parse relevant data for the decision-making algorithms, a 3D point cloud can be projected onto a discretized 2D surface where each discretization has a ternary value of either occupied, unoccupied or unobservable. This is known as an occupancy grid. In particular, the decision-making algorithm model predictive control (MPC) maintains a bird's-eye view of the world, and could interface with an occupancy grid.

\begin{figure}[t]
  \centering
  \includegraphics[width=\columnwidth]{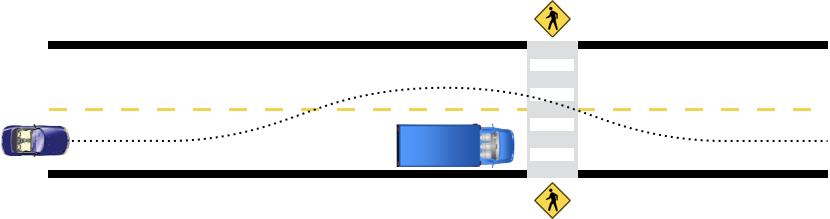}
  \vspace{-15pt}
  \caption{Example scenario of occluded pedestrian crosswalk.}
  \label{fig:scene}
\end{figure}

This work focuses on the scenario of a pedestrian crosswalk on a two-lane roadway with a large vehicle occluding the event of a pedestrian crossing as shown in Fig.~\ref{fig:scene}, which mimics Fremont Road on the Stanford campus. The underlying decision-making algorithm for path tracking is the deterministic MPC problem as presented by Brown, Funke, Erlien and Gerdes \cite{Brown2016}. Only one path option is available to the autonomous vehicle and the trajectory paths around the occluding vehicle. However, the speed along the path is only upper bounded by the speed limit. In order for the autonomous vehicle to navigate the scenario safely, it likely needs to reduce its speed. Here it is proposed that the calculation of this speed scale factor be computed through a partially observable Markov decision process (POMDP). A POMDP is like an MDP; however, there is also state uncertainty. In this scenario, the uncertainty is around the event of a pedestrian crossing. Since solving POMDPs is difficult due to the continuous state space of a belief-state, the solution is approximated using a fully observable value approximation technique called QMDP \cite{Kochenderfer2015}. 

\section{Related Work}\label{sec:related}
Motion planning under uncertainty is a large topic of study in the robotics community. For an autonomous vehicle, motion planning encompasses both lateral and longitudinal motion which are controlled through a combination of steering and acceleration. A sampled-based motion planning technique known as Rapidly-Exploring Random Trees (RRT) has been largely a deterministic planning algorithm paired with closed-loop stable control \cite{Kuwata2008}. RRT was extended to include uncertainty about the environment through chance-constraints \cite{Luders2010}, but this inclusion still depends on the knowledge that an obstacle is there. More recently, the RRT framework was expanded to branching through the belief space rather than state space to capture state uncertainty \cite{Bry2011}. RRTs are generally paired with a closed-loop controller because the planner aspect of RRT takes more than one control loop to solve for a new path. While this approach could potentially pair with MPC, the added complexity of two loops with a complicated tree search does not seem beneficial for this application of just scaling the desired speed to be lower and hence make the vehicle proceed more cautiously. 

Partially observable Markov decision processes have been used to solve subsets of larger control problems. For example, collision avoidance for unmanned aircraft was shown to perform very well in the midst of state uncertainty due to the POMDP problem formulation \cite{Temizer2010}. In general, POMDPs can be very difficult to solve because the belief space can be extremely large and intractable. However, using a QMDP solver allows the POMDP to be more closely related to an MDP \cite{Wei2011}. 

A more general form of POMDPs is a predictive state representation (PSR). PSRs represent the dynamics of a system by keeping track of probabilities for future events conditioned on previous events. It has been shown to be at least as compact as POMDPs and to maintain more information about the environment than a POMDP because it is expressed solely as observations by the agent. A PSR captures observations based only on actions taken. However, PSRs can be even more difficult to solve because they are susceptible to poor models \cite{Singh2004}. A slight variation of PSRs, called transformed PSR (TPSR), has led to successful closed-loop control by just observing actions taken by an agent in an unknown environment \cite{Boots2011}. In general, PSRs are even more difficult to solve for than POMDPs because of the need for extensive history and test events to build a good dynamic model.

\section{Infrastructure}\label{sec:sim}
For this work, the simulation environment is two-fold. The first part is a nonlinear vehicle dynamic model that simulates vehicle pose information and how the vehicle maneuvers in space. The other aspect is the simulation of the occupancy grid. 

\subsection{Vehicle Motion} 

A vehicle is controlled by commanding a steer angle ($\delta$) and longitudinal acceleration ($a_{\text{x}}$). The vehicle motion is simulated using a lumped axle vehicle model, where the two front tires are lumped as one front tire and the two rear tires are lumped as one rear tire (also known as a bicycle or single-track model). Inputs to the vehicle model are tire forces, which are modeled using a nonlinear brush tire model. Using the brush tire model, front lateral tire force is nonlinearly related to the control input steer angle ($\delta$). 

The longitudinal acceleration ($a_{\text{x}}$) control input is directly related to front and rear longitudinal tire forces by the mass of the vehicle ($\sum F_{\text{x}} = m a_{\text{x}}$). Note that most vehicles have only one driven axle but can proportionally brake on each axle. The simulation models a vehicle with front wheel drive and a brake distribution of \SI{70}{\%} in the front and \SI{30}{\%} in the rear, which means the rear longitudinal tire force is non-positive.

The full nonlinear vehicle state $x$ is 
\begin{equation}
  x = [U_{\text{y}} \quad r \quad U_{\text{x}} \quad \Psi \quad N \quad E \quad s \quad e]^\top
\end{equation}
\noindent where $U_{\text{y}}$, $r$ and $U_{\text{x}}$ are lateral velocity, yaw rate and longitudinal velocity, respectively, in the body-fixed frame; $\Psi$, $N$ and $E$ are heading, distance North and distance East, respectively, in the inertial (global) frame; and $s$ and $e$ are respectively the distance along the path and lateral deviation from the path in the path-coordinate frame. More information on how the nonlinear vehicle state is linearized and used in the MPC problem can be found in Brown et al. \cite{Brown2016}.

\subsection{Occupancy Grid} 

The occupancy grid is a discretized top-down view of the world around the vehicle. To emulate the Velodyne HDL-32E lidar, the longitudinal range of the occupancy grid is limited to \SI{70}{\meter} in front of the vehicle. Because the width of the roadway in the scenario is much less than the range of the lidar, the lateral range of the occupancy grid is limited to 8 m on each side of the vehicle. There are three tiles for every meter within range. Thus, the dimensions of the occupancy grid is 210 $\times$ 48 with the vehicle centered at (0, 24). 
 
\begin{figure}[t]
  \centering
  \includegraphics[width=\columnwidth]{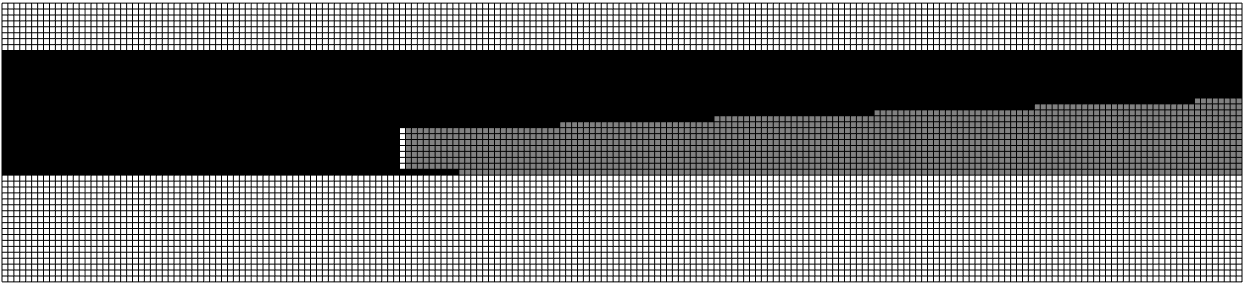}
  \vspace{-10pt}
  \caption{Example of occupancy grid simulation. Black tiles represent drivable areas, white tiles suggest occupied spaces, and gray tiles denote unknown.}
  \label{fig:hmSim}
\end{figure}

Simulation of the occupancy grid returns the number of unobservable tiles to be used as an observation in the POMDP. The simulation is a rough approximation of what is available in real-time on our test vehicle because the simulation aligns with the road heading, while the real-time occupancy grid aligns with the vehicle heading. The discrepancy can be considered minor if the curvature of the trajectory around the occluding vehicle is kept small. An instance of the occupancy grid is shown in Fig.~\ref{fig:hmSim}.

\section{Problem Formulation}\label{sec:prob}

\subsection{POMDP} 
A POMDP makes decisions based on the history of observations $o_{1:t}$. To reduce the information stored, the history is summarized in a belief state $b$, which is a distribution over states. The optimal policy is represented as a set of alpha vectors, which convert the belief state to an action. A POMDP model takes a similar form to an MDP model with the addition of observations. 

\subsubsection{State Space} 
The state space is represented in a low dimensional subspace that captures pose and motion of the vehicle as well as perception information. The components of the state considered in this work are:
\begin{itemize}
  \item V: current velocity of the vehicle
  \item D: distance along the path
  \item C: event of pedestrian crossing
\end{itemize}
Speed and distance along the path are continuous states. To further reduce the problem size, states V and S are discretized. The max speed considered for the scenario is \SI{10}{\meter\per\second}, so the speed discretization is set to intervals of \SI{1}{\meter\per\second}. The distance along the path is discretized into \SI{0.5}{\meter} intervals for a path that is \SI{60}{\meter} long. State C is already discretized as a binary occurrence.

\subsubsection{Action Space} 
The vehicle actuation considered here is longitudinal acceleration. Commanded longitudinal acceleration is determined by proportional speed control. Thus, the POMDP action space is a speed scaling factor applied to the desired speed in the longitudinal control. After discretization of the action space, the actions are $\mathcal{A}$ = \{\SI{0}{\%}, \SI{10}{\%}, \SI{20}{\%}, \SI{30}{\%}, \SI{40}{\%}, \SI{50}{\%}, \SI{60}{\%}, \SI{70}{\%}, \SI{80}{\%}, \SI{90}{\%}, \SI{100}{\%}\}. 
 
\subsubsection{Observation Space} 
The observation space captures information the agent observes after taking an action. In this work, the observations are provided just from the lidar sensors. Two types of observations are considered:
\begin{itemize}
  \item N: number of unobservable tiles
  \item C: detection of pedestrian crossing
\end{itemize}
To simplify the problem size, the number of unobservable tiles is reduced to 10 discretized bins linearly spaced between 0 and 1800 unobservable tiles. The detection of a pedestrian crossing is handled by a different perception algorithm specifically designed to detect pedestrians. For example, it could be an image recognition algorithm using cameras.

\subsubsection{Reward Model} 
The reward function in this POMDP formulation is designed with the following objectives in mind:
\begin{itemize}
  \item Encourage the vehicle to drive to the end of the path.
  \item If a pedestrian is detected, then the vehicle should yield to the pedestrian. Thus, non-zero scale factors are penalized when a pedestrian crossing event is true. 
  \item Additionally, the vehicle should not drive fast when it cannot see the pedestrian. 
\end{itemize}
To achieve the goals outlined, the reward function is specified using two costs:
\begin{itemize}
  \item {\bf Complete path reward:} The reward for the vehicle to drive to the end of the path +100.
  \item {\bf Not yielding cost:} The cost when the vehicle does not yield to the crosswalk -50.
  \item {\bf Too fast cost:} The cost to deter the vehicle from speeding around the occlusion because it is close to a pedestrian crosswalk is set to -5, which is orders of magnitude lower than the collision cost. For this work, it is simply implemented as penalizing the vehicle for going faster than \SI{6}{\meter\per\second} when it cannot see a pedestrian crossing.
\end{itemize}
The reward is assumed to be zero for all other states.

\subsubsection{State-Transition Model} 
The dynamics of the system are not actually stochastic, but rather uncertainty is introduced from the crude discretization of the state space. Also, the event of a pedestrian crossing is modeled as a random process. The following parameters characterize the state-transition model:
\begin{itemize}
  \item Speed scaling and changes in speed are not immediately realized in the state space discretization because the vehicle simulation is closer to continuous time.
  \item Vehicle is assumed to be stopped or moving forward (does not reverse direction).
  \item If a pedestrian is present in the crosswalk, the person is assumed to be standing still.
\end{itemize}
Even though the state in a POMDP is a belief state, the state transition function for a POMDP is the same as for an MDP (assumes no state uncertainty). Even though the state includes truth about a pedestrian crossing event, the problem only maintains a belief about whether there is a pedestrian crossing event based on observations.

\subsubsection{Observation Model} 
In a typical POMDP problem, the observation model is defined as the conditional probability of observing each observation $o$ given the current state $s$ and the action $a$ taken to get there: Pr$(o|s,a)$. For this work, it is assumed the action does not contribute to the observation $o$ given $s$. Thus, the dependence on $a$ is dropped and the observation model need only specify Pr$(o|s)$. Using the observation space described above, the observation model is the probability of having a number of unobservable tiles and detecting a pedestrian given the current state Pr$(o_n, o_c|s)$. A simple observation model of uniform distribution is implemented if the state where a pedestrian is not crossing is observed. Given a state where the pedestrian is crossing, the observation distribution increases to favor detecting a pedestrian crossing by \SI{30}{\%}.

\subsection{QMDP} 
QMDP is an offline method to approximate an optimal POMDP solution, and assumes at the next time step the state will be fully observable. QMDP is well suited for this problem because the actions are not information gathering. The algorithm is akin to value iteration for MDP, except it iterates over the belief state. The solver is available in the POMDPs.jl package \cite{Egorov2017}. The policy is exported as a set of alpha vectors, where each alpha vector corresponds to an action. The approximately optimal action taken is then
\begin{equation}
  \arg \max_a \boldsymbol{\alpha}^\top_a \mathbf{b}
\end{equation}
where $\alpha_a$ is an alpha vector for action $a$ and $\mathbf{b}$ is the belief state as a vector.

\begin{algorithm}[t]
\caption{POMDP policy execution}\label{alg:policy}
\begin{algorithmic}[1]
\Function{POMDPPolicyExecution}{$\boldsymbol{\alpha}$}
   \State $t \leftarrow 0$ 
   \State $b_0 \leftarrow$ uniform distribution
   \Loop
      \State Execute action $a_t = \arg \max_a \boldsymbol{\alpha}^\top_a \mathbf{b}_t$
      \State Observe $o_t$
      \State $b_{t+1}(s_{t+1}) \propto O(o_t|s_{t+1}) \sum_{s_t} T(s_{t+1}|s_t,a_t)b_t(s_t)$ 
      \State Normalize $b_{t+1}$
      \State $t \leftarrow t + 1$
   \EndLoop
\EndFunction
\end{algorithmic}
\end{algorithm}

\subsection{Policy Execution}
The execution of the POMDP policy follows Algorithm~\ref{alg:policy}. The belief state is initialized to a uniform distribution. Using the current belief state, an optimal action is calculated using the set of alpha vectors and is executed in the vehicle longitudinal control. In the next control loop (not represented in Algorithm~\ref{alg:policy}), the sensor information updates, and the perception algorithms form an observation about the number of unobservable tiles and whether a pedestrian is detected. The action, observation and belief state update the belief state using a discrete state filter.

\section{Evaluation Framework}\label{sec:eval} 
To evaluate the POMDP approach for speed control in an uncertain environment, an oracle approach and baseline approach are also implemented.

\subsection{Oracle} 
The oracle approach does not use unobservable information from the occupancy grid. Instead, perfect sensing is assumed and exact pose information about the scenario is used to allow the autonomous vehicle to maneuver around the parked vehicle at the speed limit since it will also know with perfect knowledge whether a pedestrian is about to cross the street or not. This is achievable in a physical experiment by having the pedestrian wear a GPS/INS unit as the pedestrian traverses the roadway so the autonomous vehicle has knowledge about where he or she is walking at all times. This is similarly accomplished in the simulation environment where the MPC problem will know if a pedestrian is in the crosswalk even if it would not have ``seen" the pedestrian through the obstructing vehicle. 

\subsection{Baseline}  
The baseline speed scale factor is determined by a simple function inversely proportional to the number of unobservable tiles. The speed scale factor is discretized into 10 bins, where a scale factor of 1.0 means the number of unobservable spaces is negligible and the vehicle can maneuver at the desired speed and a scale factor of 0.0 means the vehicle should come to a stop because the uncertainty in the environment is too great.

\section{Simulation Results}\label{sec:results}
\begin{figure}[t]
  \centering
  \subfloat[Hidden pedestrian]{
    \includegraphics[width=\columnwidth]{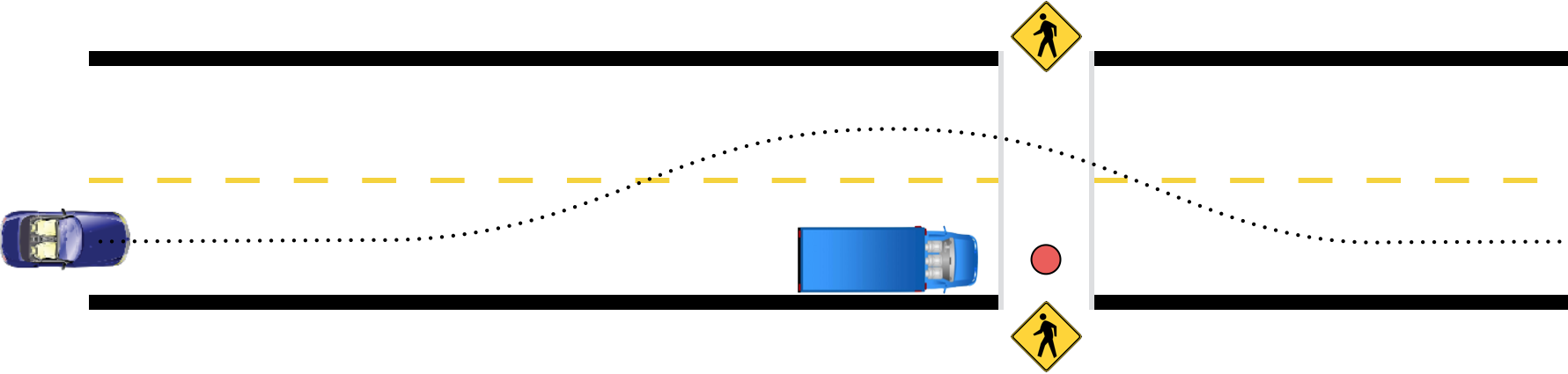}
    \label{fig:exa}}\\
  \subfloat[Exposed pedestrian]{
    \includegraphics[width=\columnwidth]{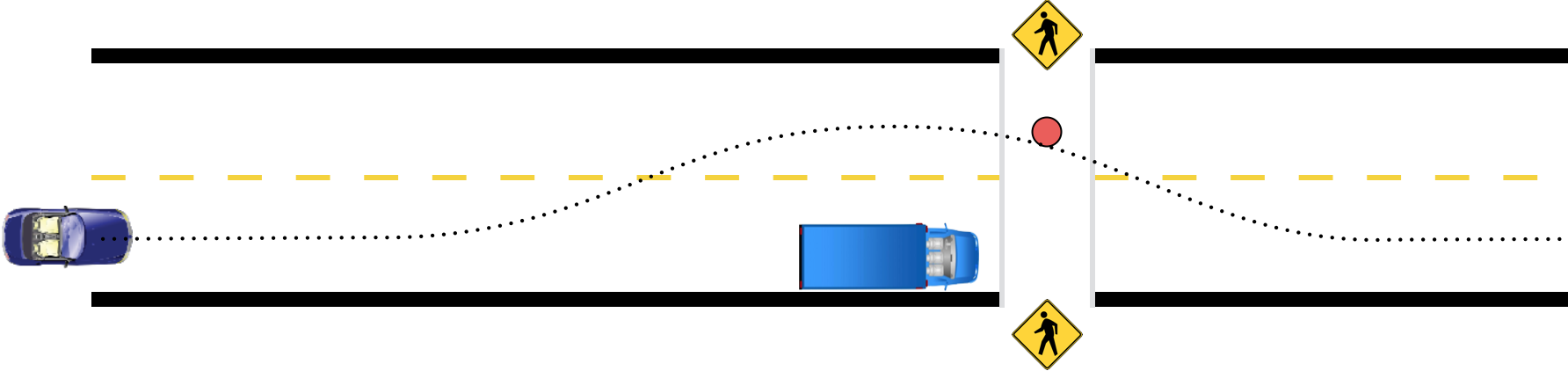}
    \label{fig:exb}}
  \caption{Simulation scenarios with pedestrian position represented by red dot.}
  \label{fig:ex}
\end{figure}

The examples for comparison are a pedestrian hidden behind the occluded vehicle in the pedestrian crosswalk (Fig.~\ref{fig:exa}), and a pedestrian in the crosswalk not behind the occluding vehicle (Fig.~\ref{fig:exb}). Since the pedestrian is in the crosswalk, all vehicles must yield. However, if an approach is unable to account for a pedestrian in the crosswalk when it is occluded, then the vehicle may perform an illegal maneuver by not yielding to the pedestrian. 

\subsection{Hidden Pedestrian} 

\subsubsection{Oracle} 
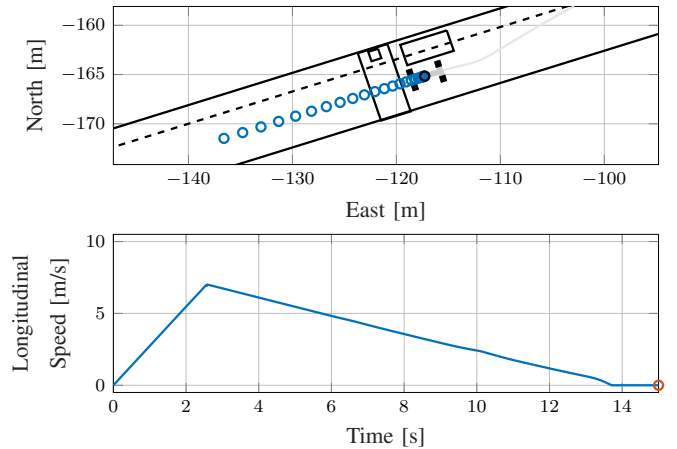
\begin{figure}
  \newlength\fwidth
  \setlength\fwidth{\columnwidth}
  \resizebox{\columnwidth}{!}{
%
%
\definecolor{mycolor1}{rgb}{0.85000,0.32500,0.09800}%
\definecolor{mycolor2}{rgb}{0.00000,0.44700,0.74100}%
\definecolor{mycolor3}{rgb}{0.92900,0.69400,0.12500}%
\definecolor{mycolor4}{rgb}{0.49400,0.18400,0.55600}%
\definecolor{mycolor5}{rgb}{0.46600,0.67400,0.18800}%
\definecolor{mycolor6}{rgb}{0.30100,0.74500,0.93300}%
\definecolor{mycolor7}{rgb}{0.63500,0.07800,0.18400}%
\begin{tikzpicture}[font=\footnotesize]

\begin{axis}[%
width=0.951\fwidth,
height=0.276\fwidth,
scale only axis,
xmin=-147.221152029511,
xmax=-94.7845267418231,
xlabel style={font=\color{white!15!black}},
xlabel={East [m]},
ymin=-174.137247042163,
ymax=-158.088340966656,
ylabel style={font=\color{white!15!black}},
ylabel={North [m]},
axis background/.style={fill=white},
xmajorgrids,
ymajorgrids,
name=plotA
]
\addplot [color=black, line width=1.0pt, forget plot]
  table[row sep=crcr]{%
-115.155845034182	-160.522015698495\\
-114.479103491934	-162.589053648073\\
-118.952458723503	-164.05361568916\\
-119.629200265751	-161.986577739582\\
-115.155845034182	-160.522015698495\\
};
\addplot [color=black, line width=1.0pt, forget plot]
  table[row sep=crcr]{%
-121.7573764716	-162.384981712578\\
-121.446230934934	-163.335343988246\\
-122.396593210602	-163.646489524912\\
-122.707738747268	-162.696127249244\\
-121.7573764716	-162.384981712578\\
};
\addplot [color=black, line width=1.0pt, forget plot]
  table[row sep=crcr]{%
-120.86422911849	-161.899079036256\\
-118.623981254495	-168.741687421066\\
-121.475068081499	-169.675124031063\\
-123.715315945494	-162.832515646254\\
-120.86422911849	-161.899079036256\\
};
\addplot [color=black, dashed, line width=1.0pt, forget plot]
  table[row sep=crcr]{%
-103.453751490275	-158.030346501373\\
-147.408006739921	-172.420827572172\\
};
\addplot [color=black, line width=1.0pt, forget plot]
  table[row sep=crcr]{%
-94.6458485247687	-160.828711264986\\
-135.511426378494	-174.207969341621\\
};
\addplot [color=black, line width=1.0pt, forget plot]
  table[row sep=crcr]{%
-109.240805972448	-158.030994856833\\
-147.255296999169	-170.47681632347\\
};
\addplot [color=white!90!black, line width=1.0pt, forget plot]
  table[row sep=crcr]{%
-102.90145865825	-158.08142169195\\
-103.251510302678	-158.271741588764\\
-103.669308828459	-158.506671549059\\
-104.18701011115	-158.80576806434\\
-104.867747990471	-159.207063396733\\
-105.762977502181	-159.74254345156\\
-106.553472410311	-160.222501352084\\
-107.154268933288	-160.594338741031\\
-107.672427618294	-160.92221912479\\
-109.640804826525	-162.179733829969\\
-110.556847485903	-162.759514513725\\
-110.898783390066	-162.968006726967\\
-111.163738357851	-163.122473551645\\
-111.402580015784	-163.254687778075\\
-111.639775134991	-163.378688636753\\
-111.852239079568	-163.482690507098\\
-112.064219792407	-163.578800714849\\
-112.252361528191	-163.657328000075\\
-112.440286569221	-163.729155878042\\
-112.628033455257	-163.794018788919\\
-112.815581106613	-163.85172883913\\
-113.025578350287	-163.908789637264\\
-113.410023283352	-164.004280677558\\
-114.24763594932	-164.211798757165\\
-114.558855623318	-164.295676082463\\
-114.850176490555	-164.38075872881\\
-115.215549040643	-164.494788442249\\
-116.188411107611	-164.808435208768\\
-117.278808136649	-165.161938869288\\
-117.278891842473	-165.161678740789\\
};
\addplot [color=mycolor2, line width=1.0pt, draw=none, mark size=2.0pt, mark=o, mark options={solid, mycolor2}, forget plot]
  table[row sep=crcr]{%
-117.278891842473	-165.161678740789\\
-117.309517145761	-165.171779958854\\
-117.34015091159	-165.181855328952\\
-117.370802018061	-165.191877733826\\
-117.401466297757	-165.201859902432\\
-117.432138876269	-165.21181672315\\
-117.46281576505	-165.221760378593\\
-117.493493860109	-165.231700349582\\
-117.524170965825	-165.241643342422\\
-117.554845810193	-165.251593242313\\
-117.585517438316	-165.261552965898\\
-117.709014749357	-165.301672586552\\
-117.89013329319	-165.360572422375\\
-118.175308472284	-165.453357721176\\
-118.564521396058	-165.580076512053\\
-119.056300456289	-165.745125943802\\
-119.653203067691	-165.940569939885\\
-120.354040334447	-166.169888198255\\
-121.158710827459	-166.433211958021\\
-122.067163633481	-166.730491314022\\
-123.079327903879	-167.061710465144\\
-124.195127611284	-167.426842711998\\
-125.414478167201	-167.825860866523\\
-126.737287319962	-168.258734595805\\
-128.163454937969	-168.725431178641\\
-129.692873114328	-169.225915288388\\
-131.320969694038	-169.758691109073\\
-133.025463428546	-170.316466897636\\
-134.784952924915	-170.892239329404\\
-136.584032034786	-171.480966944146\\
};
\addplot [color=black, line width=1.0pt, draw=none, mark size=2.0pt, mark=o, mark options={solid, black}, forget plot]
  table[row sep=crcr]{%
-117.278891842473	-165.161678740789\\
};
\addplot [color=white!80!black, line width=2.5pt, forget plot]
  table[row sep=crcr]{%
-117.278891842473	-165.161678740789\\
-118.460079005356	-165.541654432434\\
};
\addplot [color=white!80!black, line width=2.5pt, forget plot]
  table[row sep=crcr]{%
-117.278891842473	-165.161678740789\\
-115.749859910637	-164.669804985211\\
};
\addplot [color=white!80!black, line width=2.5pt, forget plot]
  table[row sep=crcr]{%
-118.216622626521	-166.29845955372\\
-118.703535384191	-164.784849311148\\
};
\addplot [color=white!80!black, line width=2.5pt, forget plot]
  table[row sep=crcr]{%
-115.506403531803	-165.426610106497\\
-115.993316289472	-163.912999863925\\
};
\addplot [color=black, line width=3.0pt, forget plot]
  table[row sep=crcr]{%
-119.007057802171	-164.886210752453\\
-118.40001296621	-164.683487869843\\
};
\addplot [color=black, line width=3.0pt, forget plot]
  table[row sep=crcr]{%
-116.297942250241	-164.010994884336\\
-115.688690328703	-163.815004843514\\
};
\addplot [color=black, line width=3.0pt, forget plot]
  table[row sep=crcr]{%
-118.520145044501	-166.399820995025\\
-117.913100208541	-166.197098112415\\
};
\addplot [color=black, line width=3.0pt, forget plot]
  table[row sep=crcr]{%
-115.811029492572	-165.524605126908\\
-115.201777571033	-165.328615086086\\
};
\end{axis}

\begin{axis}[%
width=0.951\fwidth,
height=0.276\fwidth,
scale only axis,
xmin=0,
xmax=15,
xlabel style={font=\color{white!15!black}},
xlabel={Time [s]},
ymin=-0.5,
ymax=10.5,
ylabel style={font=\color{white!15!black}, align=center},
ylabel={Longitudinal\\[1ex]Speed [m/s]},
axis background/.style={fill=white},
xmajorgrids,
ymajorgrids,
at=(plotA.below south west), anchor=above north west,
]
\addplot [color=mycolor2, line width=1.0pt, forget plot]
  table[row sep=crcr]{%
0	0\\
1.93	5.275798997259\\
2.54	6.93721705615618\\
2.55	6.93096840378669\\
2.56	6.95816758795917\\
2.57	6.95185952105014\\
2.58	6.97905770478601\\
2.59	6.97268967899593\\
2.6	6.99988712430045\\
4.32	5.89672859015579\\
5.79	4.95888826755325\\
6.19	4.70703059090782\\
6.41	4.57598070293732\\
6.48	4.53485938359978\\
6.65	4.42907318739228\\
7.3	4.0036162057409\\
7.7	3.75203943106843\\
8.48	3.26798490913483\\
9.04	2.92598423701344\\
9.27	2.7894675565619\\
9.45	2.6879664441828\\
9.59	2.61284725182053\\
9.84	2.49058452511153\\
10.07	2.38720849726041\\
10.14	2.34698708976382\\
10.19	2.31284948536234\\
10.61	2.01068073976126\\
10.8	1.8827263833043\\
10.94	1.79337383062325\\
11.11	1.69035403694384\\
11.46	1.48488061283433\\
11.91	1.22536852882101\\
12.27	1.02259755742618\\
12.63	0.825715216357358\\
12.88	0.693462735924355\\
13.06	0.59804337536449\\
13.14	0.549845699232479\\
13.22	0.496809706087078\\
13.3	0.438550726496436\\
13.36	0.388976115098703\\
13.42	0.333430749562737\\
13.48	0.272270292503944\\
13.55	0.194660318485345\\
13.64	0.0873407437012137\\
13.71	0.000579016706968361\\
13.75	0\\
15	0\\
};
\addplot [color=mycolor1, line width=1.0pt, draw=none, mark size=2.0pt, mark=o, mark options={solid, mycolor1}, forget plot]
  table[row sep=crcr]{%
15	0\\
};
\end{axis}

\end{tikzpicture}%
}
  \vspace{-15pt}
  \caption{Overhead view at \SI{15}{\second} (top) and vehicle speed (bottom) for the oracle approach with hidden pedestrian. Red circle in time series data corresponds to point in time of overhead view.}
  \label{fig:oracle_hidped}
\end{figure}

Since the oracle approach has perfect sensing, it is immediately aware of the pedestrian hidden behind the occlusion. With this information, the vehicle is able to yield to the pedestrian while attempting to maneuver around the obstruction. The trajectory of the vehicle, its final position after a \SI{15}{\second} simulation, and the vehicle speed during the maneuver are shown in Fig.~\ref{fig:oracle_hidped}. The blue circles in the overhead view represent the prediction horizon from MPC. The obstacle across the roadway represents the pedestrian crossing event, and the prediction horizon illustrates the vehicle plans to accelerate back up to speed once the crossing event is over. Additional information is needed to clear the crossing event for the oracle approach; for example, some form of communication with the pedestrian that they do not intend to cross at this time.

\subsubsection{Baseline}  
\begin{figure}
  \setlength\fwidth{\columnwidth}
  \resizebox{\columnwidth}{!}{
    \input{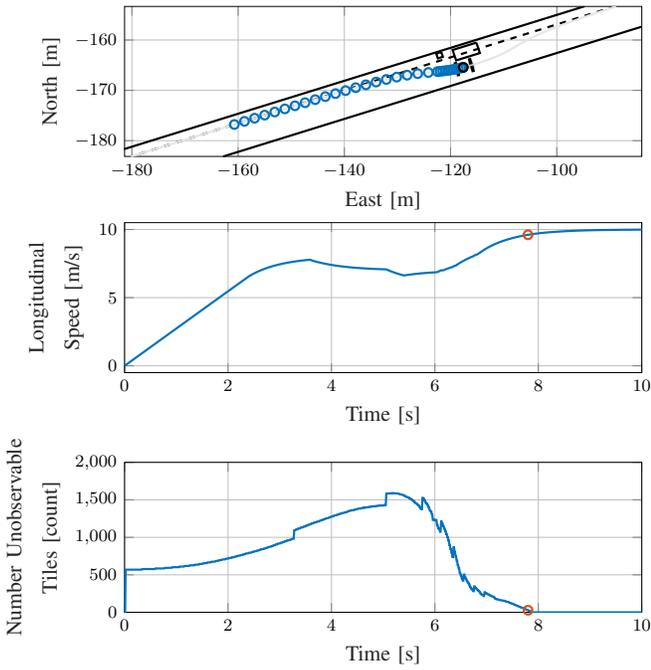}}
  \vspace{-15pt}
  \caption{Overhead view (top), vehicle speed (middle), and number of unobservable tiles (bottom) for baseline approach with hidden pedestrian. Red circles in time series data corresponds to point in time of overhead view.}
  \label{fig:baseline_hidped}
\end{figure}

Using information from the occupancy grid, the baseline approach slows the vehicle down to navigate around the occlusion. However, since it does not anticipate that a pedestrian could be hidden behind the occlusion, the vehicle goes back up to speed, narrowly missing the pedestrian. Figure~\ref{fig:baseline_hidped} shows the baseline maneuver at \SI{7.8}{\second} when the vehicle passes the pedestrian, and the speed of the vehicle. One could tune the number of bins as well as the bounds of the linearized regime for the number of unobservable tiles, but it would likely over-fit to this specific scenario and is not easily generalizable. The number of unobservable tiles recorded throughout the baseline maneuver are also depicted in Fig.~\ref{fig:baseline_hidped}.

\subsubsection{POMDP} 
\begin{figure}
  \setlength\fwidth{\columnwidth}
  \resizebox{\columnwidth}{!}{
    \input{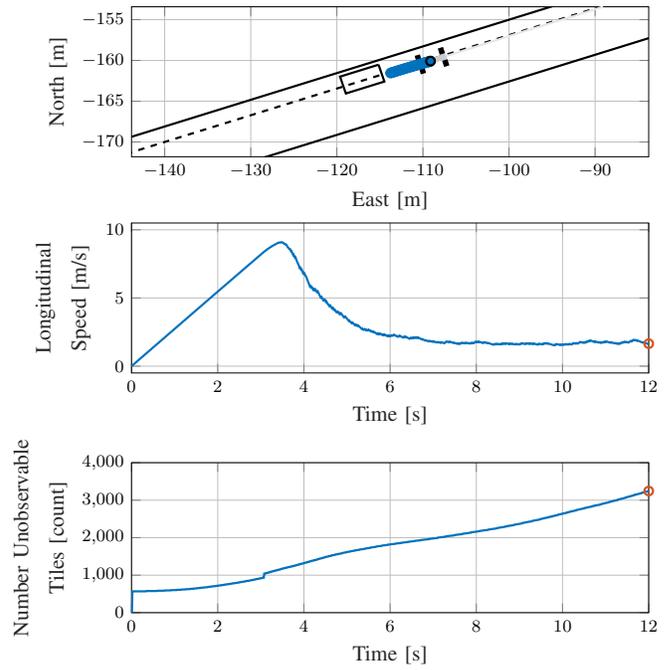}}
  \vspace{-15pt}
  \caption{Overhead view (top), vehicle speed (middle), and number of unobservable tiles (bottom) for POMDP policy with hidden pedestrian. Red circles in time series data corresponds to point in time of overhead view.}
  \label{fig:pomdp_hidped}
\end{figure}

For the POMDP approach, the QMDP policy is incorporated into the control architecture to scale the desired speed based on the observation of the action just taken and is used to determine the current belief state. At the end of a \SI{12}{\second} simulation, the vehicle is slowly approaching the back of the occluding vehicle (Fig.~\ref{fig:pomdp_hidped}). The simulation is stopped at this point because the vehicle has gotten too close to maneuver around the occluding vehicle. The prediction horizon is a function of speed. Thus, when the speed scaling reduced the vehicle speed to approximately \SI{2}{\meter\per\second} it reduced the length of the prediction horizon and was unable to plan a path around the occluding vehicle. In combination with the uncertainty created by the presence of a pedestrian in the crosswalk, the vehicle was unable to successfully maneuver around the occluding vehicle. Essentially, the autonomous vehicle is being too conservative.

\subsection{Exposed Pedestrian} 

\subsubsection{Oracle} 
\begin{figure}
  \setlength\fwidth{\columnwidth}
  \resizebox{\columnwidth}{!}{
%
%
\definecolor{mycolor1}{rgb}{0.85000,0.32500,0.09800}%
\definecolor{mycolor2}{rgb}{0.00000,0.44700,0.74100}%
\definecolor{mycolor3}{rgb}{0.92900,0.69400,0.12500}%
\definecolor{mycolor4}{rgb}{0.49400,0.18400,0.55600}%
\definecolor{mycolor5}{rgb}{0.46600,0.67400,0.18800}%
\definecolor{mycolor6}{rgb}{0.30100,0.74500,0.93300}%
\definecolor{mycolor7}{rgb}{0.63500,0.07800,0.18400}%
\begin{tikzpicture}[font=\footnotesize]

\begin{axis}[%
width=0.951\fwidth,
height=0.276\fwidth,
scale only axis,
xmin=-151.167579997176,
xmax=-94.5393465031949,
xlabel style={font=\color{white!15!black}},
xlabel={East [m]},
ymin=-174.842331538342,
ymax=-157.510529661685,
ylabel style={font=\color{white!15!black}},
ylabel={North [m]},
axis background/.style={fill=white},
xmajorgrids,
ymajorgrids,
name=plotA
]
\addplot [color=black, line width=1.0pt, forget plot]
  table[row sep=crcr]{%
-115.155845034182	-160.522015698495\\
-114.479103491934	-162.589053648073\\
-118.952458723503	-164.05361568916\\
-119.629200265751	-161.986577739582\\
-115.155845034182	-160.522015698495\\
};
\addplot [color=black, line width=1.0pt, forget plot]
  table[row sep=crcr]{%
-120.154885093526	-167.279628022223\\
-119.84373955686	-168.229990297891\\
-120.794101832528	-168.541135834557\\
-121.105247369194	-167.590773558889\\
-120.154885093526	-167.279628022223\\
};
\addplot [color=black, line width=1.0pt, forget plot]
  table[row sep=crcr]{%
-120.86422911849	-161.899079036256\\
-118.623981254495	-168.741687421066\\
-121.475068081499	-169.675124031063\\
-123.715315945494	-162.832515646254\\
-120.86422911849	-161.899079036256\\
};
\addplot [color=black, dashed, line width=1.0pt, forget plot]
  table[row sep=crcr]{%
-101.790617507856	-157.485841812208\\
-151.209455842593	-173.665409718836\\
};
\addplot [color=black, line width=1.0pt, forget plot]
  table[row sep=crcr]{%
-94.4082579558517	-160.75092488082\\
-137.649741498747	-174.908046799119\\
};
\addplot [color=black, line width=1.0pt, forget plot]
  table[row sep=crcr]{%
-107.577671990029	-157.486490167668\\
-151.294336670758	-171.7991848543\\
};
\addplot [color=white!90!black, line width=1.0pt, forget plot]
  table[row sep=crcr]{%
-101.669819398768	-157.496836141769\\
-101.966604088095	-157.623559670918\\
-102.216448994434	-157.737279182616\\
-102.502525660919	-157.875619728813\\
-102.7829402128	-158.018867307676\\
-103.096909807306	-158.186800944674\\
-103.480692951912	-158.399727114432\\
-103.966930153213	-158.677784460327\\
-104.58399142379	-159.038993125321\\
-105.457669987962	-159.559089839486\\
-106.326502460109	-160.083806500067\\
-106.998355579768	-160.497087848992\\
-107.552177912156	-160.845440516702\\
-108.172784961011	-161.244681732175\\
-108.771392809858	-161.627151247639\\
-110.679684245186	-162.835357340986\\
-110.995449275932	-163.025207394457\\
-111.259480887686	-163.176329743839\\
-111.497642350915	-163.305294593367\\
-111.734278439948	-163.425860923305\\
-111.946502313544	-163.526400044323\\
-112.158320276848	-163.618889533919\\
-112.346347329631	-163.694097007296\\
-112.534182488162	-163.762471379042\\
-112.721835542533	-163.823776565986\\
-112.909183093096	-163.87796897218\\
-113.163639884427	-163.943663002069\\
-114.487475747377	-164.275822426883\\
-114.800138875965	-164.365699862117\\
-115.154253576351	-164.475291892808\\
-115.979466418212	-164.740862694936\\
-117.278808136649	-165.161938869288\\
-117.278891842473	-165.161678740789\\
};
\addplot [color=mycolor2, line width=1.0pt, draw=none, mark size=2.0pt, mark=o, mark options={solid, mycolor2}, forget plot]
  table[row sep=crcr]{%
-117.278891842473	-165.161678740789\\
-117.309517145761	-165.171779958854\\
-117.34015091159	-165.181855328952\\
-117.370802018061	-165.191877733826\\
-117.401466297757	-165.201859902432\\
-117.432138876269	-165.21181672315\\
-117.46281576505	-165.221760378593\\
-117.493493860109	-165.231700349582\\
-117.524170965825	-165.241643342422\\
-117.554845810193	-165.251593242313\\
-117.585517438316	-165.261552965898\\
-117.709014749357	-165.301672586552\\
-117.89013329319	-165.360572422375\\
-118.175308472284	-165.453357721176\\
-118.564521396058	-165.580076512053\\
-119.056300456289	-165.745125943802\\
-119.653203067691	-165.940569939885\\
-120.354040334447	-166.169888198255\\
-121.158710827459	-166.433211958021\\
-122.067163633481	-166.730491314022\\
-123.079327903879	-167.061710465144\\
-124.195127611284	-167.426842711998\\
-125.414478167201	-167.825860866523\\
-126.737287319962	-168.258734595805\\
-128.163454937969	-168.725431178641\\
-129.692873114328	-169.225915288388\\
-131.320969694038	-169.758691109073\\
-133.025463428546	-170.316466897636\\
-134.784952924915	-170.892239329404\\
-136.584032034786	-171.480966944146\\
};
\addplot [color=black, line width=1.0pt, draw=none, mark size=2.0pt, mark=o, mark options={solid, black}, forget plot]
  table[row sep=crcr]{%
-117.278891842473	-165.161678740789\\
};
\addplot [color=white!80!black, line width=2.5pt, forget plot]
  table[row sep=crcr]{%
-117.278891842473	-165.161678740789\\
-118.460079005356	-165.541654432434\\
};
\addplot [color=white!80!black, line width=2.5pt, forget plot]
  table[row sep=crcr]{%
-117.278891842473	-165.161678740789\\
-115.749859910637	-164.669804985211\\
};
\addplot [color=white!80!black, line width=2.5pt, forget plot]
  table[row sep=crcr]{%
-118.216622626521	-166.29845955372\\
-118.703535384191	-164.784849311148\\
};
\addplot [color=white!80!black, line width=2.5pt, forget plot]
  table[row sep=crcr]{%
-115.506403531803	-165.426610106497\\
-115.993316289472	-163.912999863925\\
};
\addplot [color=black, line width=3.0pt, forget plot]
  table[row sep=crcr]{%
-119.007057802171	-164.886210752453\\
-118.40001296621	-164.683487869843\\
};
\addplot [color=black, line width=3.0pt, forget plot]
  table[row sep=crcr]{%
-116.297942250241	-164.010994884336\\
-115.688690328703	-163.815004843514\\
};
\addplot [color=black, line width=3.0pt, forget plot]
  table[row sep=crcr]{%
-118.520145044501	-166.399820995025\\
-117.913100208541	-166.197098112415\\
};
\addplot [color=black, line width=3.0pt, forget plot]
  table[row sep=crcr]{%
-115.811029492572	-165.524605126908\\
-115.201777571033	-165.328615086086\\
};
\end{axis}

\begin{axis}[%
width=0.951\fwidth,
height=0.276\fwidth,
at=(plotA.below south west), anchor=above north west,
scale only axis,
xmin=0,
xmax=15,
xlabel style={font=\color{white!15!black}},
xlabel={Time [s]},
ymin=-0.5,
ymax=10.5,
ylabel style={font=\color{white!15!black}, align=center},
ylabel={Longitudinal\\[1ex]Speed [m/s]},
axis background/.style={fill=white},
xmajorgrids,
ymajorgrids
]
\addplot [color=mycolor2, line width=1.0pt, forget plot]
  table[row sep=crcr]{%
0	0\\
1.93	5.275798997259\\
2.54	6.93721705615618\\
2.55	6.93096840378669\\
2.56	6.95816758795917\\
2.57	6.95185952105014\\
2.58	6.97905770478601\\
2.59	6.97268967899593\\
2.6	6.99988712430045\\
4.32	5.89672859015579\\
5.79	4.95888826755325\\
6.19	4.70703059090782\\
6.41	4.57598070293732\\
6.48	4.53485938359978\\
6.65	4.42907318739228\\
7.3	4.0036162057409\\
7.7	3.75203943106843\\
8.48	3.26798490913483\\
9.04	2.92598423701344\\
9.27	2.7894675565619\\
9.45	2.6879664441828\\
9.59	2.61284725182053\\
9.84	2.49058452511153\\
10.07	2.38720849726041\\
10.14	2.34698708976382\\
10.19	2.31284948536234\\
10.61	2.01068073976126\\
10.8	1.8827263833043\\
10.94	1.79337383062325\\
11.11	1.69035403694384\\
11.46	1.48488061283433\\
11.91	1.22536852882101\\
12.27	1.02259755742618\\
12.63	0.825715216357358\\
12.88	0.693462735924355\\
13.06	0.59804337536449\\
13.14	0.549845699232479\\
13.22	0.496809706087078\\
13.3	0.438550726496436\\
13.36	0.388976115098703\\
13.42	0.333430749562737\\
13.48	0.272270292503944\\
13.55	0.194660318485345\\
13.64	0.0873407437012137\\
13.71	0.000579016706968361\\
13.75	0\\
15	0\\
};
\addplot [color=mycolor1, line width=1.0pt, draw=none, mark size=2.0pt, mark=o, mark options={solid, mycolor1}, forget plot]
  table[row sep=crcr]{%
15	0\\
};
\end{axis}

\end{tikzpicture}%
}
  \vspace{-15pt}
  \caption{Overhead view at \SI{15}{\second} (top) and vehicle speed (bottom) for the oracle approach with exposed pedestrian. Red circle in time series data corresponds to point in time of overhead view.}
  \label{fig:oracle_seeped}
\end{figure}
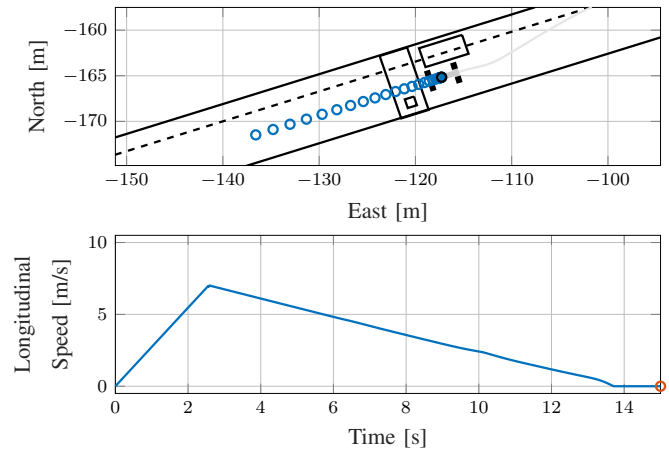

Since the oracle approach has perfect sensing, the results are the exact same as in the previous scenario. Figure~\ref{fig:oracle_seeped} depicts the overhead view and vehicle speed for the oracle approach when the pedestrian is not hidden.

\subsubsection{Baseline}  
\begin{figure}
  \setlength\fwidth{\columnwidth}
  \resizebox{\columnwidth}{!}{
    \input{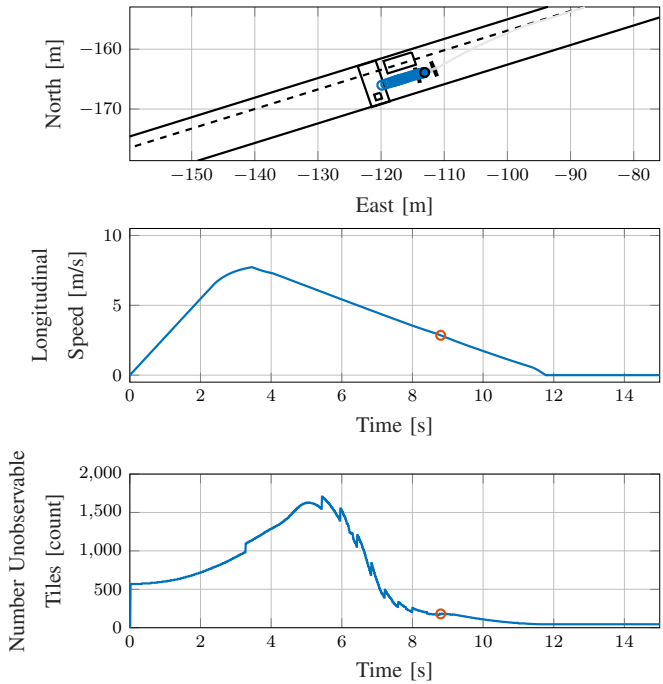}}
  \vspace{-15pt}
  \caption{Overhead view (top), vehicle speed (middle), and number of unobservable tiles (bottom) for baseline approach with exposed pedestrian. Red circles in time series data corresponds to point in time of overhead view.}
  \label{fig:baseline_seeped}
\end{figure}

Using information from the occupancy grid, the baseline approach slows the vehicle to allow it to navigate around the occlusion. For this scenario, the pedestrian is detected at the start. Thus, the ending position (Fig.~\ref{fig:baseline_seeped}) is similar to that of the oracle approach as seen in Fig.~\ref{fig:oracle_seeped}, although there is a difference in the speed transient of the vehicle compared to the previous scenario and that of the oracle approach for this scenario. The speed starts being reduced like in the previous scenario, which is shown in Fig.~\ref{fig:baseline_seeped}. But then the prediction horizon detects the crosswalk and the vehicle successfully comes to a complete stop.

\subsubsection{POMDP} 
\begin{figure}
  \setlength\fwidth{\columnwidth}
  \resizebox{\columnwidth}{!}{
    \input{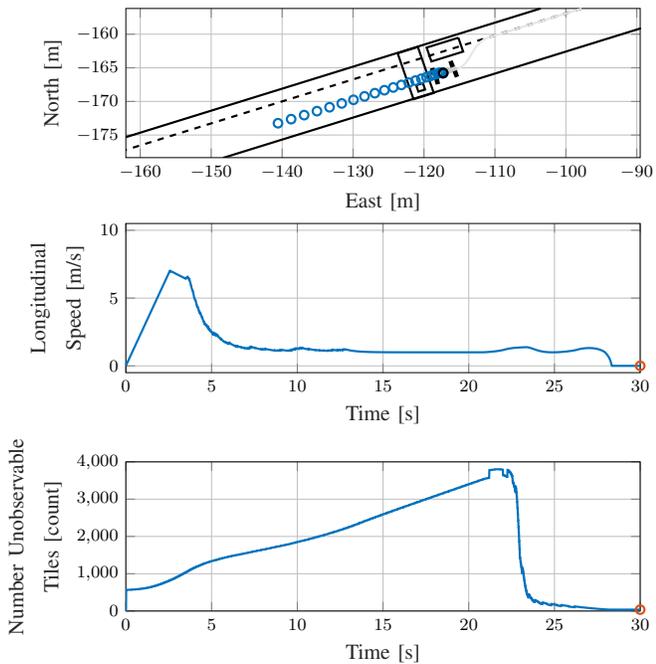}}
  \vspace{-15pt}
  \caption{Overhead view (top), vehicle speed (middle), and number of unobservable tiles (bottom) for POMDP policy with exposed pedestrian. Red circles in time series data corresponds to point in time of overhead view.}
  \label{fig:pomdp_seeped}
\end{figure}

Using the POMDP policy, dramatically different behavior is captured in Fig.~\ref{fig:pomdp_seeped}. The vehicle barely manages to maneuver around the occluding vehicle. Also, the vehicle speed plot depicts more variability towards the end of the maneuver until the vehicle successfully comes to a stop at the crosswalk. 

\section{Conclusions}\label{sec:conclusion} 
The preliminary results presented above demonstrate how well the baseline and oracle approaches work. Next steps are to keep improving the POMDP problem formulation. In particular, the observation model needs to be improved. Running many simulations with the pedestrian randomly placed throughout the crosswalk should allow us to build a more robust observation model. 

Also, next steps include thinking about how to reformulate the problem statement to make it more general to other driving scenarios with considerable uncertainty. Some generality can be incorporated by further limiting the search space for the number of unobservable tiles to just the autonomous vehicle's lane if the scenario has more than one moving vehicle. And the problem could be framed to consider whether it is safe to proceed to the end of the prediction horizon rather than to the strict location of a crosswalk.

\section*{ACKNOWLEDGMENTS}

The author thanks Mykel Kochenderfer for the many discussions regarding POMDPs.

\printbibliography

\end{document}